\documentclass[conference]{IEEEtran}
\IEEEoverridecommandlockouts
\usepackage{cite}
\usepackage{amsmath,amssymb,amsfonts}
\usepackage{algorithmic}
\usepackage{graphicx}
\usepackage{textcomp}
\usepackage{hyperref}
\usepackage{xcolor}
\def\BibTeX{{\rm B\kern-.05em{\sc i\kern-.025em b}\kern-.08em
    T\kern-.1667em\lower.7ex\hbox{E}\kern-.125emX}}
\begin{document}

\title{Predicting Citi Bike Demand Evolution Using Dynamic Graphs\\
{\footnotesize Note: Names in alphabetical order. Project links in \hyperref[sec:link]{Section 5}}
}

\author{\IEEEauthorblockN{Alexander Saff}
\IEEEauthorblockA{ars355@cornell.edu}
\and
\IEEEauthorblockN{Mayur Bhandary}
\IEEEauthorblockA{mb2393@cornell.edu}
\and
\IEEEauthorblockN{Siddharth Srivastava}
\IEEEauthorblockA{ss2848@cornell.edu}
}

\maketitle

\begin{abstract}
Bike sharing systems often suffer from poor capacity management as a result of variable demand. These bike sharing systems would benefit from models to predict demand in order to moderate the number of bikes stored at each station. In this paper, we attempt to apply a graph neural network model to predict bike demand in the New York City, Citi Bike dataset.

\end{abstract}

% \begin{IEEEkeywords}
% component, formatting, style, styling, insert
% \end{IEEEkeywords}

\section{Introduction}
In recent years, last mile transportation solutions such as bicycle sharing systems have become popular in major cities throughout the United States. These solutions seek to offer low cost and efficient modes of transport in urban areas without creating a substantial carbon footprint. Despite the environmental and economic benefits of bike sharing systems, administrators have struggled to manage variable demand for bikes. Our paper attempts to apply a graph neural network to predict bike demand in the Citi Bike network. We aim to adapt an existing model for predicting vehicle traffic from Yu et al \cite{spatio}. The results of these predictions can be used to inform restocking decisions across the Citi Bike network. 

% \subsection{Maintaining the Integrity of the Specifications}

\section{Problem Formulation}
We model our problem as predicting the evolution of a dynamic graph. The Citi Bike dataset \cite{citi} can be viewed as a graph where each node is a station and the distances between the nodes are the edges. Ideally, the number of bikes at the station represents the nodes value, however, due to the limitations of this dataset, we sum the number of outbound bikes and inbound bikes instead. This sum represents the traffic that a particular station received. Station traffic changes over time with bike demand which results in a dynamic graph. There are 1475 stations and we have data for the bikes at each station in 30 minutes time steps. We aim to predict the traffic at each station to inform restocking decisions and create better rental experiences.

\section{Approach}
We use a Spatio-Temporal Graph Convolutional Network to predict the traffic of Citi Bikes (departures and arrivals) \cite{spatio}. In our architecture, we only used one spatio-temporal convolution (ST-Conv) block because using 2 led to overfitting during our training. Subsequently there is an output layer with a final temporal convolution and a fully-connected layer \cite{medium}. The ST-Conv block consists of a spatial graph convolution between two temporal convolution layers. Temporal convolution layers work like time series prediction models, which applies a function to each of the values in the window of these time steps. The output of this function is passed to a Gated Linear Unit (GLU) \cite{glu}, which selects features that are relevant for subsequent predictions. Using our Citi Bike model as context, traffic information of these bikes at one station will influence the traffic information at nearby stations (essentially "spreading" information across nodes).

% \cite{b1}

\section{Analysis and Numerical Results}

\subsection{Preprocessing}

%As you can see in Figure \ref{fig:2}, we do something.

\begin{centering}
   \begin{figure}[h]
    \includegraphics[width=0.5\textwidth]{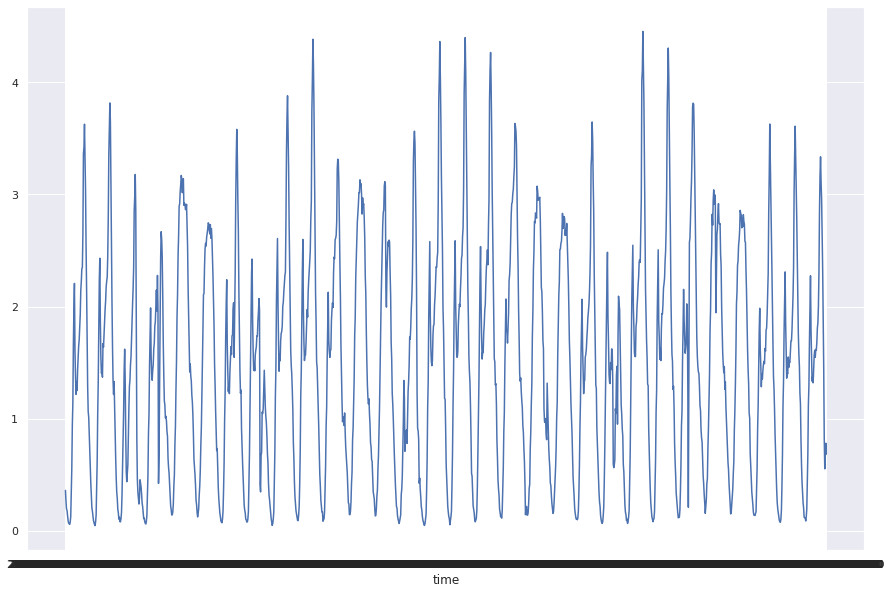}
    \caption{Average traffic for each station (inbound and outbound traffic)}
    \label{fig:1}
    \end{figure} 
\end{centering}

We focus our initial analysis on prepossessing to ensure that our data is compatible with the model that we intend to use. In Figure \ref{fig:1} we visualize the mean outbound bikes across all stations in order to understand general trends in outbound flows across the network. We can see that the data appears to be periodic and that there are generally few departures at any given time. This plot confirms that our dataset is consistent across time and that it does not include any outliers. Additionally, we suspect that the periodic spikes make this dataset suitable for inference.  
We also plotted the frequency of the average number of departures across all stations for 30 minute intervals below. This plot shows that lower departure numbers occur more frequently than higher departure numbers and that there is a peak between 1 and 2 departures during 30 minute intervals.

\subsection{Model Output and Results}
In Figure \ref{fig:1}, we show the average traffic (inbound + outbound bikes) for each station. We find that a lot of stations don't have any traffic at a give time-step, which is to be expected. 

\begin{centering}
   \begin{figure}[h]
    \includegraphics[width=0.5\textwidth]{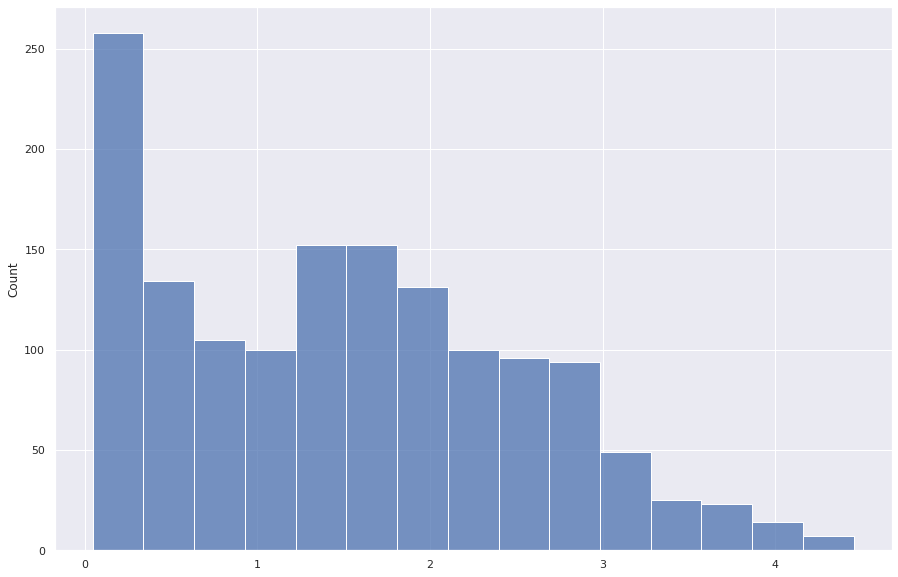}
    \caption{Average traffic across the network for a time-step (collapsed on the time-step axis)}
    \label{fig:hapd}
    \end{figure} 
\end{centering}

In Figure \ref{fig:hapd} we also plot the average traffic at all stations for one time-step. This shows the distribution of activity across the entire network, collapsed on the time-step column. Most stations don't have any traffic, but the second peak tells us that the most traffic on average at a given time-step at the stations was around one-two bikes. This basically displays two underlying distributions - a distribution when there's no traffic, and the distribution of what the "peak" traffic is on average.
% \section{Appendix and Figures}

\begin{centering}
   \begin{figure}[h]
    \includegraphics[width=0.5\textwidth]{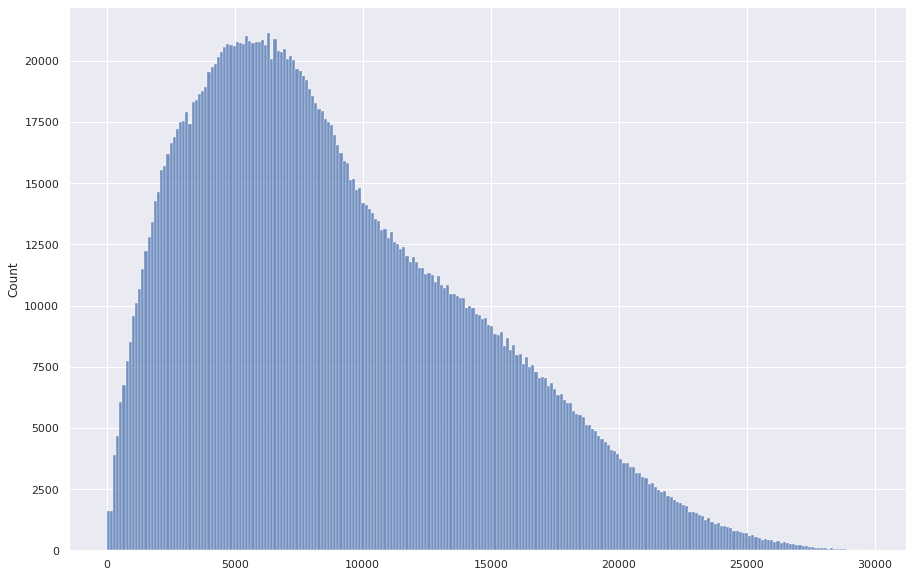}
    \caption{Pairwise distance between stations}
    \label{fig:4}
    \end{figure} 
\end{centering}

We also create a matrix of the stations' distance from each other and plot this pairwise distance in Figure \ref{fig:4}.

\begin{centering}
   \begin{figure}[h]
    \includegraphics[width=0.5\textwidth]{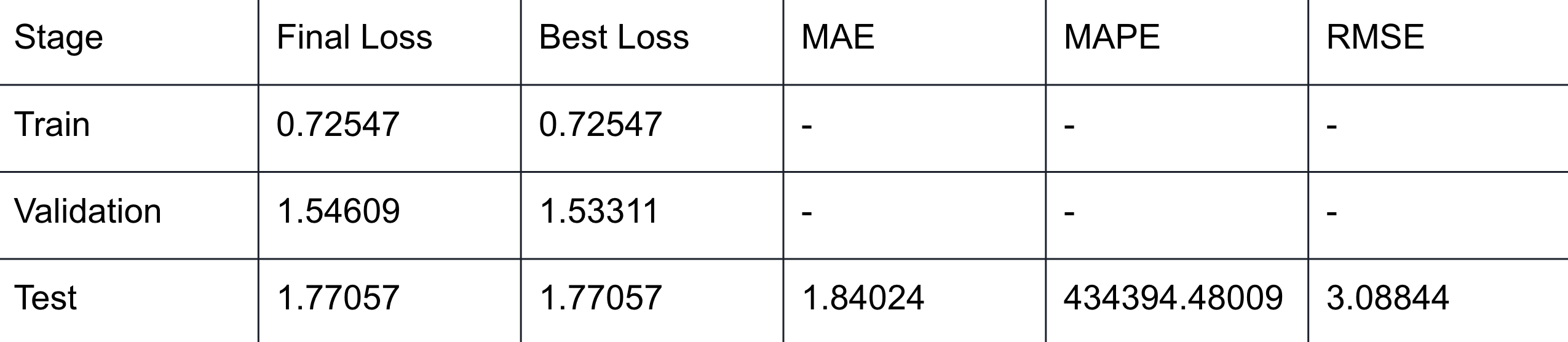}
    \caption{Final training, validation, and test results}
    \label{fig:Final_results}
    \end{figure} 
\end{centering}

The model overfit slightly, despite shrinking the number of parameters and reducing training time. This can be shown in Table \ref{fig:Final_results}, where the best validation loss is higher than the final validation loss, and the best validation loss is still higher than the best (and final) training loss. Despite this, the model performs quite well. It achieved a mean average error of 1.84, and a root mean squared error of 3.08. The mean average percent error shown in \ref{fig:Final_results} is a bad indicator of model accuracy as our ground truth data is heavily zero biased. Since the ground truth value is in the denominator of the MAPE calculation, this explodes the MAPE values.

\begin{centering}
   \begin{figure}[h]
    \includegraphics[width=0.5\textwidth]{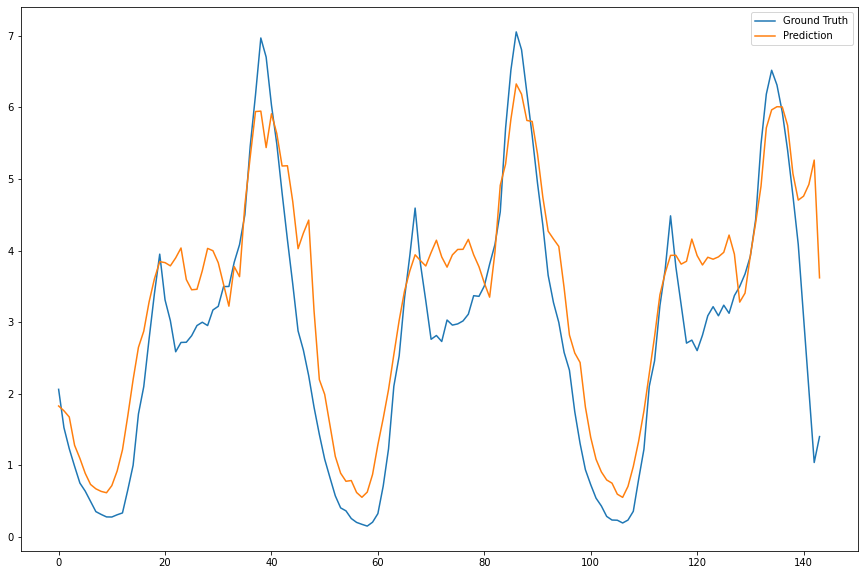}
    \caption{Prediction of our model vs Ground Truth}
    \label{fig:r}
    \end{figure} 
\end{centering}

Figure \ref{fig:r} shows the predicted average traffic (inbound + outbound bikes) for the last 3 days in June for each station overlayed on the ground-truth values. These three days were the hold-out test set, so the model never saw the ground-truth values during training. By visual inspection, the predicted values are quite close. 

\begin{centering}
   \begin{figure}[h]
    \includegraphics[width=0.5\textwidth]{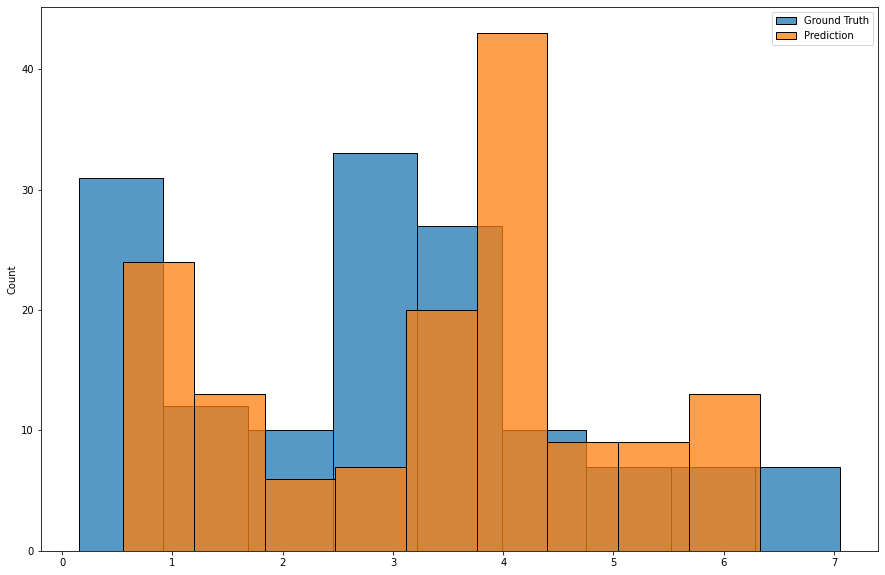}
    \caption{Model's prediction of average traffic across the entire network for a time-step}
    \label{fig:r1}
    \end{figure} 
\end{centering}

Figure \ref{fig:r1} shows the predicted average traffic at all stations for the last 3 days in June for one time-step overlayed on the ground-truth data. This shows the distribution of activity across the entire network, collapsed on the time-step column. These three days were the hold-out test set, so the model never saw the ground-truth values during training. By visual inspection, the predicted distribution is quite close.

Given more training time to find better hyperparameters, more powerful compute to operate on larger historical samples for each prediction, and a larger dataset, results could likely be improved even further.

\newpage
\section{Project Links}
\label{sec:link}
\textbf{Github (code): } \href{https://github.com/alexrsaff/GBDS_Project}{Github repo} \\

\textbf{Data: } \href{https://s3.amazonaws.com/tripdata/202106-citibike-tripdata.csv.zip}{Data} \\

\textbf{Slides: } \href{https://docs.google.com/presentation/d/1bi081WjBO9BZA_cZ3tkdIllFq8kcrh4jonUSW_fu-6w/edit?usp=sharing} {Slides} \\

% \newpage

% \begin{centering}
%   \begin{figure}[h]
%     \includegraphics[width=0.5\textwidth]{2.png}
%     \caption{Example of a parametric plot ($\sin (x), \cos(x), x$)}
%     \label{fig:2}
%     \end{figure} 
% \end{centering}

% \begin{centering}
%   \begin{figure}[h]
%     \includegraphics[width=0.5\textwidth]{3.png}
%     \caption{Example of a parametric plot ($\sin (x), \cos(x), x$)}
%     \label{fig:3}
%     \end{figure} 
% \end{centering}

% \begin{centering}
%   \begin{figure}[h]
%     \includegraphics[width=0.5\textwidth]{r2.png}
%     \caption{}
%     \label{fig:r2}
%     \end{figure} 
% \end{centering}

%\newpage 

\vspace{12pt}

\end{document}